\documentclass{arxiv}
\usepackage{booktabs}
\usepackage{graphicx}
\usepackage{multirow}
\usepackage{caption}
\usepackage{tikz}
\usepackage{xspace}
\usepackage[percent]{overpic}
\usepackage{bold-extra}

\usepackage{caption} 

\DeclareRobustCommand\onedot{\futurelet\@let@token\@onedot}
\def\onedot{. }

 \def\vs{\emph{vs}\onedot}
 
\def\vs{\emph{vs}\onedot}

\DeclareRobustCommand{\MamaDino}{\textsc{MamaDino}\xspace}

\definecolor{lightgray}{gray}{0.7} 

\title[\MamaDino: Hybrid Vision Model for Mammographic Risk Prediction]{{\fontfamily{cmr}\fontseries{bx}\fontshape{sc}\selectfont MamaDino}: A Hybrid Vision Model for Breast Cancer 3-Year Risk Prediction}

\midlauthor{\Name{Ruggiero Santeramo\nametag{$^{1}$}} \orcid{0000-0001-7107-235X} \Email{ruggiero.santeramo@fht.org}\AND
\Name{Igor Zubarev\nametag{$^{1}$}} \Email{igor.zubarev@fht.org}
\AND
\Name{Florian Jug\nametag{$^{1}$}} \Email{florian.jug@fht.org}\\
\addr $^{1}$ Fondazione Human Technopole, Milan, Italy}

\begin{document}

\maketitle

\begin{center}
\small
\textit{This manuscript extends our peer-reviewed publication at the 29th International Conference on Medical Image Computing and Computer-Assisted Intervention~(MICCAI 2026)~\cite{santeramo2026mamadino} by providing a more detailed description of the proposed architecture, and cohort characteristics, together with extended subgroup evaluations and additional analyses.}

\medskip

\textit{The peer-reviewed conference version is available at}\\
\href{https://doi.org/10.1007/978-3-032-38172-9_40}
{\texttt{https://doi.org/10.1007/978-3-032-38172-9\_40}}
\end{center}

\begin{abstract}

Breast cancer screening programmes increasingly seek to move from one-size-fits-all interval to risk-adapted and personalized strategies. The advent of deep learning (DL) gave birth to a wave of image-based risk models, able to provide more accurate short- to medium-term risk (1-5 years), compared with traditional risk models. Existing image-based risk models, such as Mirai, achieve strong discrimination but typically rely on convolutional backbones, ultra-high-resolution inputs and relatively simple multi-view fusion, with limited explicit modelling of contralateral asymmetry. 
We hypothesised that combining complementary inductive biases (convolutional and transformer-based) with explicit contralateral asymmetry modelling would allow us to match state-of-the-art 3-year risk prediction performance even when operating on substantially lower-resolution mammograms, indicating that using less detailed images in a more structured way can recover state-of-the-art accuracy.
%
In this work, we present a Mammography-Aware Multi-view Attentional DINO-based model: \MamaDino. \MamaDino is a hybrid network that fuses frozen self-supervised DINOv3 (ViT-S) features with a trainable CNN encoder at 512×512 resolution and aggregates left-right breast information via a BilateralMixer to predict a 3-year breast cancer risk score. The $512\times512$ input resolution was selected for computational feasibility rather than as an optimal clinical resolution. We train on 53,883 women from OPTIMAM, a UK cohort, and evaluate on matched 3-year case-control cohorts: an in-distribution test set from four UK screening sites and an external out-of-distribution test set from an unseen site.
%
At breast level, \MamaDino achieved discrimination comparable to Mirai on both internal and external test sets while using $\sim13\times$ fewer input pixels. Adding the BilateralMixer yielded an AUC of $0.736$ (\vs Mirai's $0.713$) on the in-distribution test set and $0.677$ (\vs $0.676$) on the external test set, with overlapping confidence intervals. Performance was further evaluated across age, ethnicity, scanner manufacturer, tumour type, and grade.

%
These findings suggest that complementary inductive biases and explicit contralateral modelling can achieve Mirai-comparable discrimination at substantially lower input resolution, motivating further investigation of resolution-efficient, bilaterally aware risk models.
\end{abstract}


\begin{figure}[!htbp]
    \centering

    \begin{overpic}[width=1\textwidth]{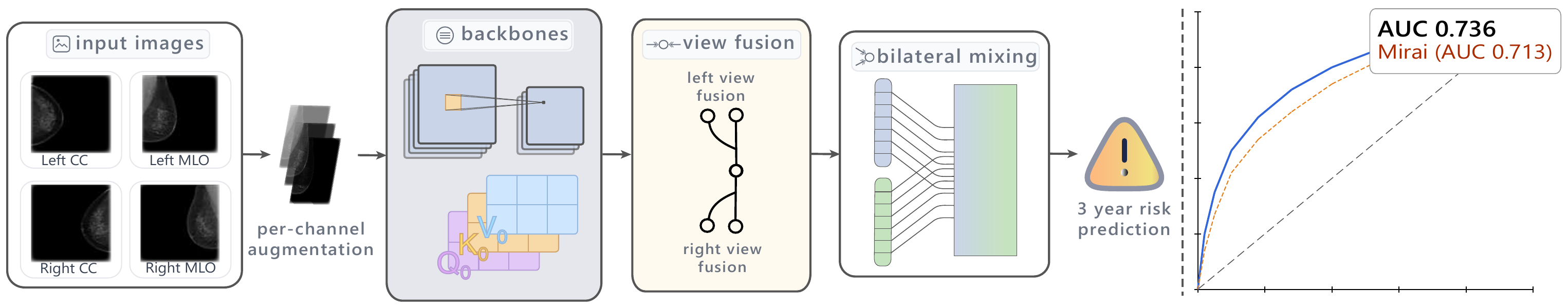}
    
    \end{overpic}
    \caption{\MamaDino overview: four standard mammography views are processed through a hybrid CNN--DINOv3 backbone with per-channel augmentation and bilateral mixing to produce a 3-year breast cancer risk score, achieving higher AUC than Mirai while operating at substantially lower image resolution.}
    \label{fig:teaser}
\end{figure}

\section{Introduction}

Breast cancer is one of the most common cancers and a leading cause of death among women worldwide, despite improvements in systemic treatment and widespread mammography screening~\cite{sung2021global}. Population screening has reduced mortality, but most programs still apply uniform interval schedules that do not reflect the large variation in individual risk. Early detection and accurate risk stratification are essential if screening is to evolve towards more personalized intervals and preventive strategies \cite{pashayan2020personalized,brentnall2023optimization}.

Risk-adapted screening has traditionally relied on clinical and demographic scores that combine age, family history, reproductive and hormonal factors to estimate long-term risk, but these tools offer only moderate discrimination and largely ignore the rich morphologic and texture information present in screening mammograms. In contrast, deep learning (DL) models that operate directly on full-field digital mammograms (FFDM), such as Mirai~\cite{Yala2021Mirai}, achieve strong short- to medium-term risk prediction and have been externally validated in multiple settings~\cite{ellis2024deep}. Recent work on the ``Better CAD, Better Risk'' paradigm further suggests that high-performing cancer detection systems can be repurposed as accurate fixed-horizon risk models, linking improved detection of early or subtle cancers to better near-term risk estimates \cite{santeramo2024better}.

From a modelling perspective, todays best-performing image-based risk models use rather high-resolution inputs (often $>$1M pixels per view) in mostly CNN-based backbones.
At the same time, self-supervised vision transformers, such as DINOv3~\cite{Simeoni2025DINOv3}, have recently emerged as powerful general-purpose encoders that have been shown to transfer well to mammography data~\cite{manigrasso2025mammography}.
The promise is that DINOv3 features can provide general purpose visual priors without requiring the training of task-specific feature generators.

This raises two related questions: can complementary CNN and transformer representations combined with explicit contralateral reasoning improve short-term risk discrimination over single-backbone baselines, and to what extent can a structured multi-view architecture maintain predictive performance at reduced input resolution?
%
%
We address these questions in the setting of 3-year fixed-horizon breast cancer risk prediction from routine screening FFDM, following the ``Better CAD, Better Risk'' view that high-quality image-based models, like DINOv3, can provide accurate risk estimates~\cite{santeramo2024better}. 

Our main hypothesis is that a resolution-efficient hybrid vision model, which combines complementary CNN and transformer inductive biases with explicit bilateral reasoning, can match or exceed the 3-year risk prediction performance of established high-resolution CNN-based approaches such as Mirai, while using substantially fewer pixels and remaining robust across imaging sites and scanner vendors.

\section{Related Work}

\paragraph{Classical and mammography-based risk models.}
Classical risk-adapted screening relies on scores such as Gail \cite{gail1989projecting}, Claus \cite{claus1991genetic} and Tyrer–Cuzick/IBIS \cite{tyrer2004breast}, which combine age, family history, reproductive and hormonal factors, and sometimes genetic information to estimate long-term risk \cite{pashayan2020personalized}. These tools do not directly exploit screening mammograms and offer only moderate discrimination. Mammographic density and parenchymal texture partially bridge this gap, with quantitative descriptors improving discrimination over purely clinical models \cite{brentnall2015mammographic}. DL-based mammography risk models operating directly on FFDMs further increase discrimination and calibration compared with Tyrer–Cuzick \cite{yala2019deep,Yala2021Mirai,omoleye2023external}, enabling imaging-enriched and image-only risk stratification.

\paragraph{Deep learning for mammographic risk prediction.}
Deep neural networks trained on FFDMs now achieve state-of-the-art short- and medium-term risk prediction. Mirai \cite{Yala2021Mirai}, a multiview model predicting five-year risk, improves upon Tyrer–Cuzick and earlier DL approaches on several settings. Subsequent work combined image-based scores with clinical covariates and revisited mammographic risk prediction with modern CNNs and larger cohorts \cite{lauritzen2023assessing,ellis2024deep}. Deep learning CAD systems trained for detection can also act as strong short-term risk predictors, suggesting a route to repurpose detection models for risk stratification \cite{santeramo2024better}. However, these systems typically use very high input resolutions and rely solely on convolutional backbones, narrowing the range of inductive biases they bring to the task, which may leave important global markers of risk under-modelled.

\paragraph{Bilateral and contralateral architectures.}
Bilateral reasoning is a natural inductive bias for paired organs such as the breasts \cite{roychoudhuri2006cancer}, and contralateral comparison is central to radiological assessment \cite{alterson2003bilateral}. Recent deep models encode this explicitly: contralateral attention improves detection over single-breast baselines \cite{mohamed2022bilateral}, while DisAsymNet \cite{wang2023disasymnet} and STA-Risk \cite{zhou2025sta} contrast left–right embeddings to capture inter-breast asymmetry for detection and risk prediction. Unlike prior single-scale CNNs with late contralateral fusion, our BilateralMixer (patient-level head) uses hybrid-encoder embeddings that decompose symmetric, asymmetric and interaction terms to capture subtle pre-diagnostic changes over three years.

\paragraph{Hybrid vision models and self-supervised transformers.}
Hybrid convolution and transformer architectures combine the local inductive biases of CNNs with the global receptive field of vision transformers, offering a favourable bias–capacity trade-off for fine-scale texture tasks compared with pure transformers \cite{Dai2021CoAtNet,Dascoli2021ConViT,goyal2022inductive}. Channel-wise gating mechanisms such as squeeze-and-excitation and Hadamard-product gating further refine this bias by reweighting feature channels based on global context \cite{hu2018squeeze,Duke2018Hadamard}. Self-supervised vision transformers like DINOv3 \cite{Simeoni2025DINOv3} learn semantically meaningful representations from large natural-image corpora and transfer well to medical imaging \cite{gao2025dino,liu2025does}, and ViT-based mammography models at reduced resolution can recover much of the performance of high-resolution CNNs at lower computational cost \cite{manigrasso2025mammography}. Most prior work, however, either fine-tunes such transformers end-to-end or uses them as single-stream backbones, without combining them with texture-specialised encoders or explicitly modelling bilateral context. Here we keep a DINOv3 branch frozen as a global prior, condition a trainable SE-ResNeXt branch via cross-attentional fusion, and couple both with a bilateral mixing head at $512\times512$ for three-year risk prediction.

\section{Methods}
\subsection{Dataset and Cohort Selection}

Data were sourced from the OPTIMAM Mammography Image Database \cite{halling2020optimam} and comprised women attending routine FFDM screening at four UK services (Jarvis, Leicester, Imperial, St George’s) between 2010 and 2021. All exams were standard two-view bilateral screenings (L-CC, R-CC, L-MLO, R-MLO) using only \textit{FOR-PRESENTATION} images. Women aged 40–80 years were included, and exams with missing views or implants were excluded. All data were anonymised.

To standardise inputs across sites and manufacturers, DICOM mammograms were converted to 8-bit PNG images.
Right-breast views were flipped to align breast orientation. Images were zero-padded on the right and bottom to preserve aspect ratios before resizing to the input resolution required by each model: $512\times512$ for \MamaDino and
$1664\times2048$ for Mirai. Pixel intensities were normalised using statistics computed from the training set.

After filtering, the training cohort comprised \textbf{53{,}883} women. Approximately $500$ developed biopsy-confirmed cancer during the screening interval (CI), $\sim1{,}500$ had benign findings (B), $\sim11{,}000$ had malignant or suspicious recalls (M), and the remainder were normal (N). 
Many women contributed multiple examinations (mean $2.57$ episodes per patient).
Dataset partitions were constructed at patient level, ensuring that all examinations from the same woman belonged exclusively to the training,
validation, or test partition.
Supervision at \emph{episode level} used the most severe bilateral outcome, while lesion-side OPTIMAM annotations provided breast-level labels. During training, episodes with CI/M/B outcomes and their screening exams in the preceding 3 years (CIP, MP) were labelled positive at both episode and breast level using lesion laterality, while normal (N) episodes and contralateral normal breasts in unilateral M episodes were labelled negative.

Imaging was predominantly acquired on Hologic systems ($\sim96\%$), with the remaining $\sim4\%$ from GE, Siemens and Philips devices. Self-reported ethnicity was available at the patient level but substantially missing and unevenly distributed (Tab. ~\ref{tab:ethnicity}), so it was used only for fairness assessment, not as a model input. 

\paragraph{Definition of the 3-year risk outcome.}
For a screening examination at time $t_i$, we define a patient-level binary outcome $y\in\{0,1\}$. Patients are labelled positive if they develop a biopsy-confirmed interval cancer within three years of baseline, or if a malignancy is detected at the first subsequent routine screen at the three-year follow-up. Controls have no cancer diagnosis before or at the subsequent three-year screening examination.

\paragraph{In-distribution validation cohort.}
The internal held-out cohort was a 1:2 matched case–control sample drawn from the four UK screening services used for training (Imperial, Jarvis, Leicester, St George’s), including women aged 47–73 years (UK screening age being 50-70). Matching was performed at the \textit{patient level} on screening site, age, and imaging manufacturer (Hologic, Siemens, GE Medical Systems); cases were defined as screening examinations acquired three years before a subsequently diagnosed breast malignancy (including interval cancers), and controls were women with normal outcomes at their subsequent screening. The final internal test set comprised 525 cases and 1,050 matched controls, with detailed distributions of age, self-reported ethnicity, imaging site, scanner type, cancer type (DCIS vs.\ invasive), and tumour grade reported in Tab.~\ref{tab:test_sets_description}.

\paragraph{Out-of-distribution validation cohort.}
The external cohort consisted of women from the Oxford screening service, which was not used in model development, and used the same age range (47–73 years) as the internal cohort. We constructed a 1:4 matched case–control sample, matching on age and scanner manufacturer (Hologic, GE Medical Systems); the final external test set comprised 376 cases and 1,504 matched controls. Scanner distribution strongly differed from the training and internal cohorts, with GE Medical Systems accounting for roughly one third of examinations, and racial/ethnic information was largely unavailable, precluding ethnicity-based stratification. Summary characteristics for this cohort, including cancer type and tumour grade for malignancies diagnosed three years after the baseline examination, are reported in Tab.~\ref{tab:test_sets_description}.

\subsection{Per-channel augmentation}
\label{subsec:per_channel_aug}

To exploit ImageNet-pretrained backbones with single-channel mammograms, we construct a pseudo-RGB image by applying distinct intensity transforms to a single greyscale view. Given a preprocessed mammogram $I \in \mathbb{R}^{H \times W}$, we define three channels $I^{(1)}, I^{(2)}, I^{(3)} \in \mathbb{R}^{H \times W}$. Channel $I^{(1)}$ undergoes a random brightness jitter to promote robustness to exposure, breast thickness, and global intensity scaling; $I^{(2)}$ receives an independent, random contrast jitter to mitigate vendor- and site-specific windowing while preserving relative tissue and lesion contrast; $I^{(3)}$ is processed with contrast-limited adaptive histogram equalisation (CLAHE) using relatively high clip limits and a fine grid (e.g.\ $12 \times 12$ tiles) to enhance subtle low-contrast structures such as early masses or microcalcifications.
During inference, the three-channel representation was constructed deterministically, without stochastic brightness or contrast perturbations. This ensures that validation and test predictions are not affected by random augmentation. 

We compare this scheme against naive three-channel replication across input resolutions in Fig.~\ref{fig:auc_vs_resolution_mean_range}.

\subsection{Model Architecture}
\label{subsec:model_arch}

We design a hybrid fusion network that couples a frozen DINOv3 vision transformer with a trainable SE-ResNeXt101 backbone to exploit complementary inductive biases. For each breast $b \in \{\mathrm{L}, \mathrm{R}\}$, the input consists of the two standard views (CC and MLO), stacked into a tensor
$x^{(b)} \in \mathbb{R}^{2 \times 3 \times H \times W}$.

\paragraph{Backbone encoders and cross-attentional fusion}
The transformer branch is a frozen DINOv3 ViT-S/16+ model that produces a grid of patch tokens for each view, while a view-specific SE-ResNeXt101 encoder (same architecture, untied weights) extracts convolutional texture features. SE blocks modulate channel responses to emphasise discriminative patterns. We resize DINO and ResNeXt feature maps to a common $16 \times 16$ grid, project them to a shared latent dimension $d$ (here $d = 512$) with $1 \times 1$ convolutions, and apply multi-head cross-attention in which ResNeXt tokens query DINO tokens, followed by residual feed-forward layers. This allows the convolutional stream to integrate globally coherent, semantically rich transformer features while preserving high-frequency CNN information.

\paragraph{BridgeMixer block and breast embedding.}
On top of the SE-ResNeXt backbone, a SE-based BridgeMixer projects ResNeXt features through a bottleneck to $d$ channels, concatenates this projection with the cross-attention output for each view, and compresses the result back to $d$ channels via a $1 \times 1$ convolution. This yields fused view-specific feature maps $F^{\text{view}} \in \mathbb{R}^{d \times 16 \times 16}$ for CC and MLO. We concatenate the two views along the channel dimension to obtain a joint breast representation $F^{(b)} \in \mathbb{R}^{2d \times 16 \times 16}$, apply adaptive average pooling to $2 \times 2$, and flatten the result into an $8d$-dimensional per-breast embedding.

\paragraph{Breast-level classifier and patient-level baseline.}
For each breast, the embedding is passed through a two-layer MLP with LayerNorm, GELU and dropout to produce a logit $z^{(b)}$, trained with a breast-level binary focal loss. At inference we obtain probabilities $p^{(\mathrm{L})}$ and $p^{(\mathrm{R})}$ and define patient risk as $p_{\text{patient}} = \max\bigl(p^{(\mathrm{L})}, p^{(\mathrm{R})}\bigr)$, assuming malignancy in either breast determines patient risk; this max aggregation provides the baseline for the bilateral fusion head.

\paragraph{The BilateralMixer: Patient-level prediction}
A key ingredient to the method we present is the BilateralMixer at patient level: a fusion module that combines left and right breast embeddings via a shallow transformer, a learned asymmetry gate, and symmetric feature composition to capture bilateral context, asymmetry, and concordance.

Let $\mathbf{e}_L, \mathbf{e}_R \in \mathbb{R}^{d}$ denote the left and right breast embeddings from the per-side encoder. The BilateralMixer combines them in three stages. First, \textit{relational encoding via a symmetric transformer}: the embeddings are treated as tokens in a compact transformer encoder sequence $[\text{CLS}, \mathbf{e}_L, \mathbf{e}_R]$, where a learnable classification token (CLS) aggregates bilateral context through self-attention and yields a patient-level embedding $\mathbf{c}$. In the absence of side-dependent positional encodings, the CLS-based relational representation is invariant to exchanging the breast tokens. The subsequent asymmetry gate learns side-specific weights from the ordered pair of breast embeddings.

Second, \textit{gated asymmetry weighting}: a small multilayer perceptron computes soft attention coefficients $(\alpha_L, \alpha_R)$ from the concatenation of both embeddings and their absolute difference,
\begin{equation}
    [\alpha_L, \alpha_R] = \mathrm{softmax}\!\big(f_\theta([\mathbf{e}_L, \mathbf{e}_R, |\mathbf{e}_L - \mathbf{e}_R|])\big),
\end{equation}
so that the model can emphasize the side with stronger pathological evidence while retaining bilateral context.

Third, \textit{symmetric feature composition}: the final representation concatenates four components,
\begin{equation}
    \mathbf{z} = [\mathbf{c}, \; \alpha_L\mathbf{e}_L + \alpha_R\mathbf{e}_R, \; |\mathbf{e}_L - \mathbf{e}_R|, \; \mathbf{e}_L \odot \mathbf{e}_R],
\end{equation}
where $\odot$ denotes elementwise multiplication. The absolute difference encodes order-invariant asymmetry, and the product captures bilateral concordance as a feature-wise co-activation signal.

The fused vector $\mathbf{z}$ is passed to a shallow multilayer perceptron that outputs a single patient-level malignancy logit, providing a fully differentiable analogue of the radiologist’s comparison of contralateral mammograms.
\subsection{Training and Evaluation}

\begin{figure}[!htbp]
    \centering
    \begin{overpic}[width=1\textwidth]{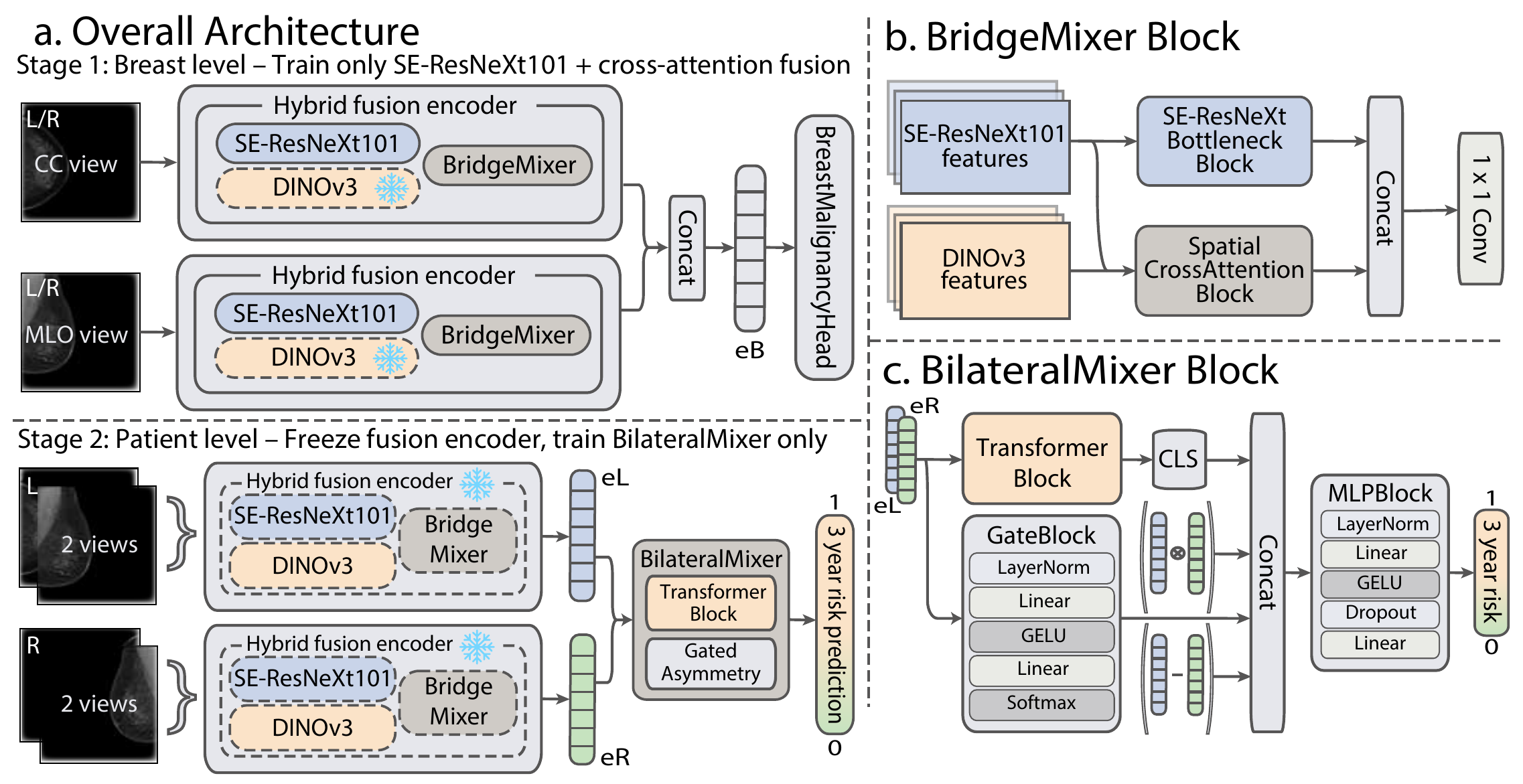}
    \end{overpic}
    \caption{
    \textbf\textbf{\MamaDino{} architecture:}
    \textbf{(a)} Four standard mammography views per exam (at \(512\times512\) resolution) are processed by a hybrid fusion encoder that combines a frozen DINOv3 for global semantics with a trainable SE-ResNeXt101 CNN for local texture, producing per-breast embeddings that are fused for 3-year malignancy risk prediction in \textbf{Stage 2} via the BilateralMixer.
    \textbf{(b)} The BridgeMixer block aligns Transformer tokens with convolutional feature maps via spatial cross-attention and 1×1-convolution fusion.
    \textbf{(c)} The BilateralMixer block takes left and right breast embeddings, models bilateral concordance and asymmetry through a transformer, and outputs the final risk score.
    }
    \label{fig:image-diagram}
\end{figure}

Training used two stages: first learning breast-level representations, then patient-level aggregation. In stage one (Fig.~\ref{fig:image-diagram}.a), we trained the hybrid fusion encoder on the large screening cohort using $512\times512$ mammograms with the per-channel and geometric/photometric augmentations (Sec.~\ref{subsec:per_channel_aug}), while validation and test images underwent only deterministic resizing per-channel augmentation and normalisation. The DINOv3 branch was initialised from a self-supervised ViT-S/16+ checkpoint on natural images and kept frozen; the SE-ResNeXt101 branch, initialised from ImageNet, was trained jointly with the cross-attention fusion modules to predict breast-level malignancy. Optimisation used AdamW with a binary focal loss to address class imbalance and early stopping based on breast-level AUC on a held-out validation subset.

In stage two (Fig.~\ref{fig:image-diagram}.a), we froze the fusion encoder and trained only the BilateralMixer (Fig.~\ref{fig:image-diagram}.c) head for patient-level prediction on the same training set, using the same optimiser and loss with early stopping on patient-level AUC. All hyperparameters (learning rates, regularisation, number of epochs) were tuned on the internal training/validation split and fixed before evaluation on the matched in-distribution and external OOD test cohorts. Both stages were trained on a single NVIDIA H200 GPU. Stage one used a batch size of 32, an initial learning rate of $1\times10^{-5}$, and up to 50 epochs. Stage two used a batch size of 64, a learning rate of $5\times10^{-7}$, and up to 25 epochs.

For comparison, we trained three baselines under the same preprocessing and optimisation pipeline: (i) a DINO-only model with a linear classifier on frozen transformer features, (ii) a SE-ResNeXt-only model trained directly on $512\times512$ mammograms, and (iii) a hybrid breast-level fusion model with max aggregation across breasts at inference (patient risk $=\max(p^{(\mathrm{L})}, p^{(\mathrm{R})})$). 
Subgroup AUCs were computed for pre-specified strata of age, self-reported ethnicity, tumour type and grade, and imaging manufacturer, and reported only where sample sizes were adequate. For the tumour-specific analysis, cancer cases were stratified according to the lesion diameter, molecular subtype, and radiological appearance of the subsequently diagnosed malignancy. These characteristics were obtained from follow-up records. AUCs were computed by comparing cancer cases belonging to that subgroup against all controls from the corresponding test cohort. The same control population was reused across subgroups, and paired DeLong tests were used to compare \MamaDino and Mirai within each subgroup.
Mirai was evaluated using its publicly released model and intended inference preprocessing. As structured clinical risk factors were unavailable in OPTIMAM, only mammographic images were provided, and Mirai's patient-level 3-year risk score was used for comparison. The comparison with Mirai therefore provides an external reference rather than a controlled architectural ablation.

\section{Results}
\label{sec:results}
\begin{table}[h]
\centering
\caption{\label{tab:overall_results}\textbf{Overall AUC for 3-year breast cancer risk prediction} on in-distribution and out-of-distribution test sets, comparing single-stream baselines, resolution ablations, alternative bilateral fusion strategies, Mirai, and \MamaDino variants at breast and patient level.
}
\resizebox{\linewidth}{!}{
\begin{tabular}{lccccc}
\toprule
 & &  \multicolumn{2}{c}{\textbf{In-Distribution Test Set}} & \multicolumn{2}{c}{\textbf{OOD Test Set}} \\
\cmidrule(lr){3-4} \cmidrule(lr){5-6}

Model & Resolution  & AUC  & (95\% CI) & AUC  & (95\% CI) \\
\midrule

DINO-only* & 512$\times$512 & 0.621  & -- & -- & -- \\
SEResNeXt-only* & 512$\times$512 & 0.668  & -- & -- & -- \\
\MamaDino (resolution ablation)*
& $224\times224$ & 0.638 & -- & -- & -- \\

\MamaDino (resolution ablation)*
& $320\times320$ & 0.649 & -- & -- & -- \\

\MamaDino$^\dagger$ (concat. ablation)
& $512\times512$ & 0.731 & -- & -- & -- \\

\midrule
{Mirai$^\dagger$} \cite{Yala2021Mirai} & 1664$\times$2048 & 0.713 & (0.686–0.740)  &  0.676 & (0.646-0.707)  \\
\MamaDino* (ours) & 512$\times$512 & 0.727 & (0.701–0.754) &  0.666 & (0.635-0.696)  \\
{\MamaDino$^\dagger$ (ours)} & 512$\times$512 & \textbf{0.736} & (0.710–0.762)  & \textbf{0.677} & (0.647-0.707)  \\
\bottomrule
\multicolumn{6}{l}{\footnotesize * Predictions were at breast-level, and patient risk corresponded to the max of the two breast scores.}\\
\multicolumn{6}{l}{\footnotesize $\dagger$ Predictions were at patient-level.}\\
\bottomrule
\end{tabular}
}
\end{table}
\subsection{Prediction of 3-year Cancer Risk}
In Tab.~\ref{tab:overall_results}, we report 3-years fixed-horizon risk discrimination results on the matched in-distribution and external OOD cohorts. On the in-distribution test set, the single-stream baselines achieved AUCs of 0.621 (DINO-only) and 0.668 (SE-ResNeXt-only), compared with 0.727 for the hybrid breast-level \MamaDino model at $512\times512$ resolution. Mirai achieved an in-distribution AUC of 0.713 and an OOD AUC of 0.676 at its native resolution of $1664\times2048$ pixels. In contrast, the breast-level \MamaDino, (using max aggregation) reached an in-distribution AUC of 0.727 and an OOD AUC of 0.666 at $512\times512$, achieving discrimination comparable to Mirai on the OOD cohort while using $\sim$13$\times$ fewer input pixels. Adding the BilateralMixer patient head further increased discrimination to an in-distribution AUC of 0.736 and an external OOD AUC of 0.677,  slightly outperforming Mirai on both cohorts. Confidence intervals for \MamaDino and Mirai overlapped on the external cohort, showing robustness to device manufacturer's shift.

\paragraph{Resolution ablation.}
We additionally evaluated the hybrid breast-level model at reduced input resolutions. The AUC was 0.638 at $224\times224$ and 0.649 at $320\times320$, compared with 0.727 at $512\times512$. These results show that input resolution remains important for discrimination within the proposed architecture. The $512\times512$ setting was selected under computational constraints, rather than being identified as an optimal clinical resolution.

\paragraph{Bilateral fusion ablation.}
To evaluate the contribution of the BilateralMixer, we replaced it with direct concatenation of the left and right breast embeddings followed by a patient-level prediction head. This baseline achieved an in-distribution AUC of 0.731, compared with 0.736 for the full BilateralMixer. The improvement of 0.005 is modest, suggesting that the benefit of the proposed architecture comes from the combined design rather than from the bilateral module alone.

\begin{table}[ht]
\centering
\small
\caption{\label{tab:subgrup_analysis}\textbf{Subgroup AUCs for 3-year breast cancer risk prediction} of \MamaDino and Mirai on in-distribution and out-of-distribution (Oxford) test cohorts, stratified by age, ethnicity, scanner, cancer type and tumour grade. Values are AUC estimates with corresponding 95\% confidence intervals; dashes indicate subgroups not represented or not evaluable.}
\resizebox{\linewidth}{!}{
\begin{tabular}{lcccc}
\toprule
Test Cohorts & \multicolumn{2}{c}{\textbf{In-Distribution Test}} & \multicolumn{2}{c}{\textbf{OOD Test Set}} \\
\cmidrule(lr){2-3} \cmidrule(lr){4-5}
& \MamaDino & Mirai & \MamaDino & Mirai \\
\midrule
\textbf{Age (y)} & & & & \\
\quad $<60$ & \textbf{0.746} {\footnotesize(0.710-0.783)} & 0.727 {\footnotesize(0.690-0.764)} & 0.668 {\footnotesize(0.626-0.710)} & \textbf{0.693} {\footnotesize(0.651-0.735)}\\
\quad $60–65$ & \textbf{0.705} {\footnotesize(0.650-0.760)} & 0.656 {\footnotesize(0.597-0.715)} & \textbf{0.675} {\footnotesize(0.618-0.732)} & 0.657 {\footnotesize(0.596-0.717)}\\
\quad $65+$ & \textbf{0.747} {\footnotesize(0.696-0.797)} & 0.743 {\footnotesize(0.691-0.794)} & \textbf{0.691} {\footnotesize(0.624-0.758)} & 0.672 {\footnotesize(0.606-0.737)}\\
\textbf{Ethnicity} & & & & \\
\quad White & 0.723 {\footnotesize(0.689-0.758)} & \textbf{0.726} {\footnotesize(0.691-0.760)} & -- & -- \\
\quad Asian & \textbf{0.801} {\footnotesize(0.723-0.880)} & 0.660 {\footnotesize(0.556-0.764)} & -- & --\\
\quad Black/African & 0.770 {\footnotesize(0.617-0.924)}  & \textbf{0.839} {\footnotesize(0.728-0.950)} &  -- & --\\
\textbf{Scanner} & & & & \\
\quad GE Med. S. & \textbf{0.725} {\footnotesize(0.555-0.895)} & 0.627 {\footnotesize(0.443-0.811)} & \textbf{0.669} {\footnotesize(0.618-0.721)} & 0.622 {\footnotesize(0.566-0.677)}\\
\quad Hologic, Inc. & \textbf{0.740} {\footnotesize(0.713-0.766)} & 0.722 {\footnotesize(0.694-0.749)} & 0.679 {\footnotesize(0.642-0.716)} & \textbf{0.705} {\footnotesize(0.669-0.741)}\\
\quad Siemens & \textbf{0.463} {\footnotesize(0.217-0.709)} & 0.457 {\footnotesize(0.222-0.691)} & -- & -- \\
\textbf{Cancer type*} & & & & \\
\quad DCIS & 0.696 {\footnotesize(0.636-0.757)} & \textbf{0.711} {\footnotesize(0.650-0.771)} & 0.659 {\footnotesize(0.586-0.731)} & \textbf{0.689} {\footnotesize(0.617-0.761)}\\
\quad Invasive & \textbf{0.748} {\footnotesize(0.718-0.778)} & 0.713 {\footnotesize(0.682-0.745)} & \textbf{0.680} {\footnotesize(0.647-0.714)} & 0.673 {\footnotesize(0.640-0.707)}\\
\textbf{Tumour grade*} & & & & \\
\quad G1 & \textbf{0.755} {\footnotesize(0.684-0.826)} & 0.686 {\footnotesize(0.613-0.758)} & \textbf{0.683} {\footnotesize(0.621-0.745)} & 0.679 {\footnotesize(0.614-0.745)}\\
\quad G2 & \textbf{0.787} {\footnotesize(0.751-0.824)} & 0.756 {\footnotesize(0.717-0.796)} & \textbf{0.695} {\footnotesize(0.645-0.745)} & 0.660 {\footnotesize(0.609-0.711)}\\
\quad G3 & \textbf{0.641} {\footnotesize(0.572-0.709)} & 0.633 {\footnotesize(0.561-0.705)} & 0.664 {\footnotesize(0.600-0.728)} & \textbf{0.686} {\footnotesize(0.623-0.749)}\\
\bottomrule
\multicolumn{5}{l}{\footnotesize * Cancer data from 3-year follow-up.}\\
\bottomrule
\end{tabular}
}
\end{table}

\subsection{Subgroup analysis}
Tab.~\ref{tab:subgrup_analysis} reports patient-level AUCs stratified by age, ethnicity, imaging manufacturer, cancer type and tumour grade. On the in-distribution test cohort, the hybrid model showed stable discrimination across age, with AUCs between 0.70 and 0.75 and consistently, though modestly, higher values than Mirai. On the out-of-distribution (OOD) test cohort, both models achieved AUCs around 0.66–0.69 across age strata: Mirai performed better in women younger than 60 years, while the hybrid model matched or exceeded Mirai in the older groups.

Across ethnic groups in the in-distribution cohort, \MamaDino and Mirai performed similarly in White women, with larger but less precisely estimated differences in Asian and Black women, where \MamaDino and Mirai respectively achieved higher AUCs. Stratification by scanner vendor showed that, on Hologic systems, which contributed the majority of training exams, \MamaDino slightly outperformed Mirai, while performance on Siemens devices was comparable. On GE scanners, \MamaDino was consistently superior to Mirai on both the in-distribution and OOD test cohorts.

When stratified by cancer type, both models showed higher discrimination for invasive cancers than for DCIS. \MamaDino improved over Mirai for invasive disease on both cohorts, with comparable performance for DCIS on the in-distribution test set and slightly worse performance on the OOD test cohort. By tumour grade, the hybrid model achieved AUCs that were generally equal to or higher than Mirai, with the largest gains for low- and intermediate-grade tumours and similar performance between models for high-grade disease.

\subsection{Tumour-Specific Analysis} \label{subsec:tumour_analysis}

To further investigate the discrimination of \MamaDino across different tumour characteristics, we stratified the evaluation by lesion diameter, molecular subtype, and radiological appearance (Tab.~\ref{tab:extended_tumour_analysis}). We compared the three-year risk scores of \MamaDino and Mirai using paired DeLong tests, reporting AUC differences ($\Delta$AUC), their 95\% confidence intervals, and two-sided $p$-values.

\paragraph{Lesion diameter.} On the in-distribution test set, \MamaDino achieved numerically higher AUCs than Mirai in three of the four lesion-size categories. The largest difference was observed for lesions measuring 11--20\,mm, where \MamaDino achieved an AUC of 0.745 compared with 0.698 for Mirai ($\Delta$AUC = +0.047, $p=0.051$). For lesions $\leq$10\,mm, both models showed lower discrimination, with AUCs of 0.687 and 0.665, respectively. On the OOD cohort, differences were generally small, although Mirai achieved a numerically higher AUC for lesions $>$30\,mm (0.763 vs.\ 0.701, $p=0.070$). These findings suggest that the relative discrimination of the two models may vary with tumour size, without establishing a statistically significant difference in any individual category.

\paragraph{Molecular subtype.} Both models showed variation in discrimination across molecular subtypes. For HR+/HER2$-$ cancers, the most represented subtype, \MamaDino achieved an AUC of 0.754 compared with 0.727 for Mirai on the in-distribution cohort, while the corresponding OOD AUCs were 0.664 and 0.657. For triple-negative cancers, \MamaDino showed numerically higher discrimination in-distribution (0.566 vs.\ 0.490, $p=0.060$), whereas Mirai achieved a higher AUC on the OOD cohort (0.708 vs.\ 0.595, $p=0.11$). However, these subgroups contained only 15 and 17 cancer cases, respectively. Other less prevalent molecular subtypes similarly had small sample sizes and wide confidence intervals, limiting the reliability of subgroup-specific conclusions.

\paragraph{Radiological appearance.} When stratifying by radiological appearance, \MamaDino achieved numerically higher AUCs for mass-associated lesions on the in-distribution cohort (0.743 vs.\ 0.720), while both models achieved an AUC of 0.679 on the OOD cohort. For lesions presenting exclusively as calcifications, discrimination was lower for both models, with AUCs of 0.691 and 0.675 in-distribution, and 0.629 and 0.651 out-of-distribution, for \MamaDino and Mirai, respectively. These results indicate that discrimination varies across radiological presentations, although the paired comparisons do not establish a significant advantage for either architecture.

\paragraph{Overall interpretation.} Across all evaluated tumour-specific subgroups, none of the paired comparisons reached the conventional significance threshold of $p<0.05$. The observed differences were generally modest, with some variation in direction between the in-distribution and OOD cohorts. Importantly, the absence of statistical significance does not establish equivalence between the models, particularly given the limited number of cancer cases in several subgroups. Overall, these analyses support broadly comparable discrimination between \MamaDino and Mirai across the evaluated tumour characteristics, while highlighting the need for larger cohorts to assess potential subtype- and presentation-specific differences.

\begin{table}[p]
\centering
\small
\caption{\label{tab:extended_tumour_analysis}
\textbf{Tumour-specific AUCs for 3-year breast cancer
risk prediction} of \MamaDino and Mirai on the
in-distribution and out-of-distribution (Oxford)
test cohorts, stratified by lesion diameter,
molecular subtype, and radiological appearance.
Values are AUC estimates with 95\% confidence
intervals, paired AUC differences, and corresponding
DeLong test $p$-values.
}
\resizebox{\linewidth}{!}{
\begin{tabular}{lccccc}
\toprule
& &
\multicolumn{2}{c}{\textbf{AUC (95\% CI)}}
& \textbf{$\Delta$AUC (95\% CI)}
& \textbf{$p$} \\
\cmidrule(lr){3-4}
\textbf{Subgroup} & \textbf{$cases$ $(n)$}
& \MamaDino & Mirai & & \\
\midrule

\multicolumn{6}{l}{
\textbf{In-Distribution Test Set}} \\

\textit{Largest lesion diameter}
& & & & & \\

\quad $\leq$10\,mm & 112
& \textbf{0.687} [0.63, 0.74]
& 0.665 [0.61, 0.72]
& +0.023 [-0.034, 0.079] & 0.43 \\

\quad 11--20\,mm & 156
& \textbf{0.745} [0.71, 0.78]
& 0.698 [0.65, 0.74]
& +0.047 [0.000, 0.094] & 0.051 \\

\quad 21--30\,mm & 82
& 0.724 [0.66, 0.78]
& \textbf{0.736} [0.68, 0.79]
& -0.012 [-0.071, 0.048] & 0.70 \\

\quad $>$30\,mm & 82
& \textbf{0.765} [0.71, 0.82]
& 0.746 [0.69, 0.81]
& +0.019 [-0.031, 0.069] & 0.45 \\

\addlinespace[3pt]
\textit{Molecular subtype}
& & & & & \\

\quad HR+/HER2$-$ & 260
& \textbf{0.754} [0.72, 0.79]
& 0.727 [0.69, 0.76]
& +0.027 [-0.010, 0.063] & 0.15 \\

\quad HR+/HER2+$^{\dagger}$ & 17
& 0.692 [0.55, 0.83]
& \textbf{0.702} [0.54, 0.86]
& -0.010 [-0.118, 0.098] & 0.86 \\

\quad HR$-$/HER2+$^{\dagger}$ & 12
& \textbf{0.810} [0.72, 0.90]
& 0.767 [0.64, 0.89]
& +0.043 [-0.121, 0.208] & 0.60 \\

\quad Triple negative$^{\dagger}$ & 15
& \textbf{0.566} [0.42, 0.71]
& 0.490 [0.34, 0.64]
& +0.075 [-0.003, 0.154] & 0.060 \\

\addlinespace[3pt]
\textit{Radiological appearance}
& & & & & \\

\quad Mass-associated & 309
& \textbf{0.743} [0.71, 0.77]
& 0.720 [0.69, 0.75]
& +0.023 [-0.011, 0.056] & 0.19 \\

\quad Calcifications only & 98
& \textbf{0.691} [0.64, 0.75]
& 0.675 [0.62, 0.73]
& +0.017 [-0.042, 0.075] & 0.58 \\

\quad Other imaging abnormalities$^{\dagger}$ & 16
& 0.624 [0.46, 0.79]
& \textbf{0.628} [0.47, 0.79]
& -0.004 [-0.081, 0.072] & 0.91 \\

\quad Other / unclassified & 102
& \textbf{0.770} [0.72, 0.82]
& 0.743 [0.69, 0.79]
& +0.027 [-0.026, 0.079] & 0.32 \\

\midrule

\multicolumn{6}{l}{
\textbf{OOD Test Set}} \\

\textit{Largest lesion diameter}
& & & & & \\

\quad $\leq$10\,mm & 78
& \textbf{0.634} [0.57, 0.70]
& 0.627 [0.56, 0.69]
& +0.007 [-0.056, 0.070] & 0.83 \\

\quad 11--20\,mm & 107
& \textbf{0.669} [0.61, 0.72]
& 0.649 [0.59, 0.70]
& +0.020 [-0.042, 0.083] & 0.52 \\

\quad 21--30\,mm & 58
& \textbf{0.709} [0.65, 0.77]
& 0.683 [0.62, 0.75]
& +0.026 [-0.045, 0.098] & 0.47 \\

\quad $>$30\,mm & 67
& 0.701 [0.64, 0.76]
& \textbf{0.763} [0.71, 0.82]
& -0.062 [-0.129, 0.005] & 0.070 \\

\addlinespace[3pt]
\textit{Molecular subtype}
& & & & & \\

\quad HR+/HER2$-$ & 202
& \textbf{0.664} [0.62, 0.70]
& 0.657 [0.62, 0.70]
& +0.007 [-0.037, 0.052] & 0.75 \\

\quad HR+/HER2+$^{\dagger}$ & 15
& \textbf{0.709} [0.57, 0.84]
& 0.673 [0.51, 0.84]
& +0.036 [-0.109, 0.181] & 0.63 \\

\quad HR$-$/HER2+$^{\dagger}$ & 6
& -- & -- & -- & -- \\

\quad Triple negative$^{\dagger}$ & 17
& 0.595 [0.44, 0.75]
& \textbf{0.708} [0.58, 0.84]
& -0.113 [-0.251, 0.025] & 0.11 \\

\addlinespace[3pt]
\textit{Radiological appearance}
& & & & & \\

\quad Mass-associated & 226
& \textbf{0.679} [0.64, 0.72]
& \textbf{0.679} [0.64, 0.72]
& +0.000 [-0.041, 0.042] & 0.99 \\

\quad Calcifications only & 62
& 0.629 [0.56, 0.69]
& \textbf{0.651} [0.58, 0.72]
& -0.022 [-0.089, 0.046] & 0.53 \\

\quad Other / unclassified & 88
& \textbf{0.708} [0.65, 0.77]
& 0.691 [0.63, 0.75]
& +0.017 [-0.056, 0.091] & 0.64 \\

\bottomrule
\end{tabular}
}

\vspace{2mm}
\begin{minipage}{\linewidth}
\footnotesize
$n$ indicates the number of cancer cases in each subgroup.
AUCs and 95\% CIs were estimated using DeLong's method.
$\Delta$AUC denotes \MamaDino minus Mirai, with paired
95\% CIs and two-sided $p$-values.
Mass-associated includes masses with or without
calcifications; calcifications only excludes masses.
Other imaging abnormalities include asymmetry or
architectural distortion without a mass or calcifications.
Other/unclassified includes clinical abnormalities
without a specific mammographic descriptor or lesion
record. $^{\dagger}$ Fewer than 30 cancer cases. Bold values indicate the numerically higher AUC within each subgroup, not statistical significance.
Subgroup counts may differ due to missing data.
\end{minipage}
\end{table}

\section{Discussion}


\MamaDino achieved 3-year breast cancer risk discrimination comparable to Mirai while using $512\times512$ mammograms, corresponding to approximately $13\times$ fewer input pixels per view. The controlled ablations provide further insight into the model's architecture. Combining frozen DINOv3 representations with a trainable CNN improved discrimination over the individual backbones, while explicit bilateral aggregation provided additional, albeit modest, improvements over maximum aggregation and direct embedding concatenation. The resolution ablations also showed that discrimination decreased at smaller input sizes, indicating that structured feature fusion does not eliminate the importance of spatial resolution.

Clinically, these results suggest that short-term mammographic risk prediction does not intrinsically require high-resolution inputs if information is organised through stronger priors. \MamaDino integrates global semantics from a frozen DINOv3 branch with fine-grained mammographic texture from a trainable SE-ResNeXt101 branch, fuses left–right embeddings via a bilateral mixing head, and uses per-channel augmentation to construct pseudo-RGB inputs from single-channel mammograms. This configuration appears sufficient to recover state-of-the-art performance at lower resolution. In the context of existing DL risk models such as Mirai and CNN-only bilateral architectures, our findings support the view that model design and fusion strategy can compensate for reduced pixel count.

The behaviour of \MamaDino relative to Mirai across in-distribution and OOD cohorts highlights robustness to scanner shifts, particularly between Hologic and GE systems, consistent with the hypothesis that self-supervised transformers provide stable domain priors. Discrimination for DCIS remained lower than for invasive cancers, echoing prior reports and suggesting that early in situ changes yield weaker or more diffuse signatures at this resolution. Ethnicity-stratified analyses and Siemens-specific results were limited by sample size, and wide confidence intervals in these strata are more consistent with data scarcity than systematic failure of the model.

The tumour-specific analysis provides additional insight into how both models discriminate future breast cancer risk across different tumour characteristics. Although \MamaDino showed numerically higher AUCs in several subgroups, including intermediate-sized lesions and HR+/HER2$-$ cancers, none of the paired comparisons reached statistical significance. Discrimination also varied by radiological appearance, with generally lower AUCs for calcifications-only presentations than for mass-associated lesions. However, the small numbers of cancers in several molecular subgroups limit the strength of these conclusions, motivating validation in larger, independently collected cohorts.

Limitations of this study include its restriction to UK screening services and a case–control design, which limits assessment of calibration under true population incidence and generalisability to settings with different screening intervals, prevalence or demographics. Siemens scanners and non-White groups were under-represented, constraining conclusions on device- and ancestry-specific performance. We also evaluated only a single 3-year horizon and did not retrain Mirai-like architectures at 512$\times$512, so we cannot fully disentangle the respective contributions of resolution and architecture.
The $512\times512$ resolution was selected for computational feasibility and should not be interpreted as an optimal resolution for clinical mammographic risk prediction. The additional resolution and concatenation ablations were evaluated only on the in-distribution cohort, without corresponding external estimates or confidence intervals. Their interpretation is therefore  restricted to the evaluated settings.

These limitations motivate future work on self-supervised pretraining directly on large-scale mammography or tomosynthesis, extension of the BilateralMixer to longitudinal and 3D data, evaluation in multinational cohorts with prospective impact assessment, and systematic study of ensembles combining \MamaDino with Mirai or other image, clinical and genomic risk scores for personalised, risk-adapted screening.

\section{Conclusion}



We introduced \MamaDino, a hybrid vision architecture that combines frozen DINOv3 features, trainable SE-ResNeXt encoders, cross-attentional fusion, and a bilateral aggregation head for 3-year breast cancer
risk prediction. At $512\times512$ resolution, \MamaDino achieved discrimination comparable to Mirai on internal and external UK screening cohorts while processing approximately $13\times$ fewer pixels per view. Our ablation experiments highlighted the
contributions of hybrid feature extraction, input resolution, and bilateral fusion. Together, these results support further investigation of resolution-efficient hybrid architectures with explicit contralateral reasoning.

These findings have broader implications for AI-enabled screening. Our findings argue that progress in mammographic risk prediction may hinge less on ever higher input resolution and more on how global context, local texture and bilateral relationships are represented and fused. \MamaDino combines self-supervised transformer priors, convolutional texture encoders and a dedicated bilateral mixing head, showing that appropriately structured architectures can recover state-of-the-art performance even when the pixel budget is constrained. This shifts the emphasis from brute-force detail towards the design of task-aware, multi-view, hybrid models.

In summary, \MamaDino shows that “using less detail in a more structured way” can recover state-of-the-art short-term risk prediction in mammography, pointing toward hybrid, and bilaterally aware architectures as a promising foundation for future AI systems for personalised breast cancer screening.

\section*{Data Availability}
The images and data used in this publication are derived from the OPTIMAM imaging database \cite{halling2020optimam}, we would like to acknowledge the OPTIMAM project team and staff at the Royal Surrey NHS Foundation Trust who developed the OPTIMAM database, Cancer Research UK which funded the creation and maintenance of the database, and Cancer Research Horizons which facilitates access to the OPTIMAM data. Database: \hyperlink{https://medphys.royalsurrey.nhs.uk/omidb/}{https://medphys.royalsurrey.nhs.uk/omidb/}

\bibliography{biblio}
\newpage
\appendix
\section{Supplementary Analyses}
\label{appendix_a}
\setcounter{table}{0}
\setcounter{figure}{0}
\renewcommand{\thetable}{A\arabic{table}}
\renewcommand*{\theHtable}{\thetable}
\renewcommand{\thefigure}{A\arabic{figure}}
\renewcommand*{\theHfigure}{\thefigure}
\begin{table}[h!]
\centering
\caption{\textbf{Ethnic composition of the training cohort.} Self-reported ethnic backgrounds among women in the UK OPTIMAM training cohort, highlighting the predominance of White participants and the large proportion of missing ethnicity records.}
\label{tab:ethnicity}
\begin{tabular}{l r}
\toprule
\textbf{Ethnicity} & \textbf{Number of Women} \\
\midrule
White      & 30{,}682 \\
Asian      & 3{,}808 \\
Black      & 1{,}851 \\
Mixed      & 601 \\
Other      & 1{,}107 \\
Missing    & 15{,}834 \\
\midrule
\textbf{Total} & 53{,}883 \\
\bottomrule
\end{tabular}
\end{table}

\begin{table}[ht]
\centering
\small
\caption{\label{tab:test_sets_description}\textbf{Characteristics of the matched internal (in-distribution) and external (out-of-distribution) case–control test cohorts.} Values are counts with column-wise percentages. Age is reported both categorically and as median (IQR). Cancer type and tumour grade refer to the malignancy diagnosed three years after the baseline screening examination. The Screening site and Scanner rows highlight the distributional shift between the different cohorts: internal data come from four services dominated by Hologic systems, whereas the external Oxford cohort includes a much higher proportion of GE scanners.}
\begin{tabular}{lcccc}
\toprule
Test Cohorts & \multicolumn{2}{c}{\textbf{In-Distribution Test Set}} & \multicolumn{2}{c}{\textbf{OOD Test Set}} \\
\cmidrule(lr){2-3} \cmidrule(lr){4-5}
& Control (n=1050) & Case (n=525) & Control (n=1504) & Case (n=376)\\
\midrule
\textbf{Age (y)} & & & & \\
\quad $<60$ & 532 (50.7\%) & 266 (50.7\%) & 728 (48.5\%) & 182 (48.4\%)\\
\quad $60–65$ & 260 (24.8\%) & 130 (24.8\%) & 419 (27.8\%) & 105 (27.9\%)\\
\quad $65+$ & 258 (24.6\%) & 129 (24.6\%) & 357 (23.8\%) & 89 (23.7\%)\\
\quad Median* & 59 (54–64)  & 59 (54–64) & 60 (55–64)  & 60 (55–64) \\
\\
\textbf{Ethnicity} & & & & \\
\quad White & 623 (59.3\%) & 306 (58.3\%) & 2 (0.1\%) & 0 (0.0\%)\\
\quad Asian & 88 (8.4\%) & 43 (8.2\%) &  & \\
\quad Black/African & 30 (2.9\%) & 18 (3.4\%) & & \\
\quad Other/Not Stated & 309 (29.4\%) & 158 (30.1\%) & 1502 (99.9\%) & 376 (100.0\%)\\
\\
\textbf{Screening site} & & & & \\
\quad Imperial & 246 (23.4\%) & 123 (23.4\%) &  & \\
\quad Jarvis Breast Centre & 280 (26.7\%) & 140 (26.7\%) & & \\
\quad Leicester & 242 (23.0\%) & 121 (23.0\%) & & \\
\quad St. George's & 282 (26.9\%) & 141 (26.9\%) & & \\
\quad Oxford &  &  & 1504 (100.0\%) & 376 (100.0\%)\\
\\
\textbf{Scanner} & & & & \\
\quad GE Medical Systems & 26 (2.5\%) & 13 (2.5\%) & 484 (32.1\%) & 121 (32.2\%)\\
\quad Hologic, Inc. & 1006 (95.8\%) & 503 (95.8\%) & 1020 (67.9\%) & 255 (67.8\%)\\
\quad Siemens & 18 (1.7\%) & 9 (1.7\%) & & \\
\\
\textbf{Cancer type**} & & & & \\
\quad DCIS &  & 102 (19.4\%) &  & 62 (16.5\%)\\
\quad Invasive &  & 400 (76.2\%) &  & 314 (83.5\%)\\
\quad Unknown &  & 23 (4.4\%) &  & 0 (0.0\%)\\
\\
\textbf{Tumour grade**} & & & & \\
\quad G1 &  & 78 (14.9\%) &  & 79 (21.0\%)\\
\quad G2 &  & 215 (41.0\%) &  & 143 (38.0\%)\\
\quad G3 &  & 92 (17.5\%) &  & 88 (23.4\%)\\
\quad N/A &  & 140 (26.7\%) &  & 66 (17.6\%)\\
\\
\bottomrule
\multicolumn{5}{l}{\footnotesize * Data are medians, with IQRs in parentheses.}\\
\multicolumn{5}{l}{\footnotesize ** Cancer data from 3-year follow-up.}\\
\bottomrule
\end{tabular}
\end{table}

\begin{figure}[htbp]
    \centering
    \includegraphics[width=.8\textwidth]{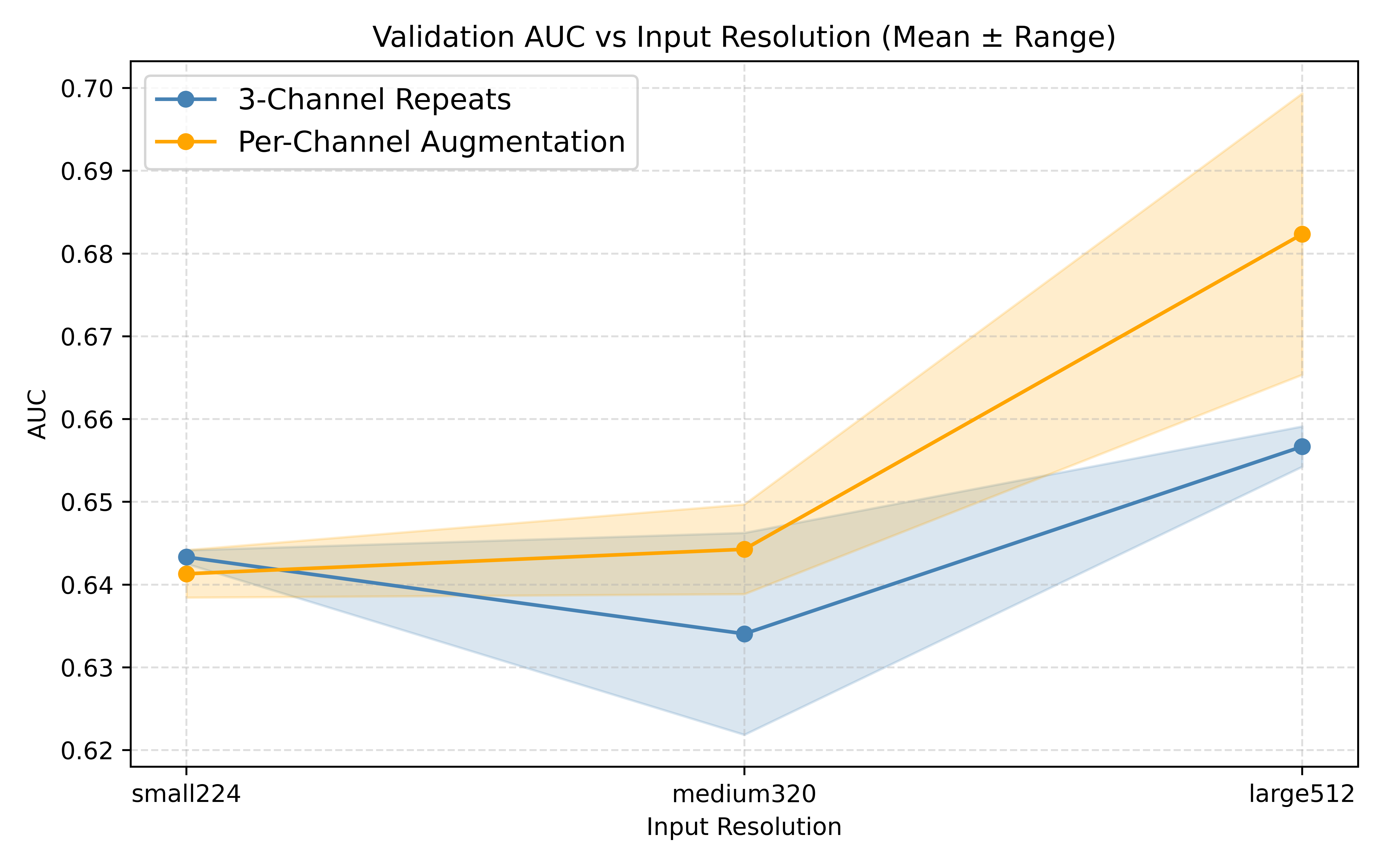}
    \caption{
    \textbf{Per-channel augmentation \vs simple replication ablation.}
    Validation AUC of the breast-level \MamaDino encoder on a held-out OPTIMAM validation set as a function of input resolution (224, 320, 512). Curves show mean AUC across random initialisations (2 seed per setup), with shaded regions indicating the range across runs. The blue line (“3-Channel Repeats”) corresponds to the standard practice of replicating the grayscale image into three identical channels before augmentation. The orange line (“Per-Channel Augmentation”) corresponds to our proposed strategy, where each channel is independently perturbed (brightness/contrast jitter and CLAHE) before being recombined. Across all resolutions, per-channel augmentation yields consistently higher AUC, with the largest gain at 512×512.}
    \label{fig:auc_vs_resolution_mean_range}
\end{figure}
\end{document}